%% file: main.tex
\documentclass[10pt,twocolumn,letterpaper]{article}

\usepackage{wacv}                       

\usepackage{microtype}
\usepackage{graphicx}
\usepackage{tikz}
\usetikzlibrary{arrows.meta,positioning}
\usepackage{booktabs}
\usepackage{float}
\usepackage{amsmath,amssymb}
\usepackage[accsupp]{axessibility}  
\usepackage{algorithm}
\usepackage{algorithmic}
\usepackage{xspace}
\usepackage{enumitem}

\definecolor{wacvblue}{rgb}{0.21,0.49,0.74}
\usepackage[pagebackref,breaklinks,colorlinks,allcolors=wacvblue]{hyperref}
\usepackage[capitalize,noabbrev]{cleveref}
\usepackage{capt-of}

\def\wacvPaperID{872} 
\def\confName{WACV}
\def\confYear{2027}

\newcommand{\netvlad}{NetVLAD\xspace}
\newcommand{\vocabtree}{vocab-tree\xspace}
\newcommand{\dinotopb}{DINOv2\xspace}
\newcommand{\wei}{Wei~et~al.\xspace}

\title{Degree-Equalized View-Graph Selection under Tight Pair Budgets}
\input{authors}

\begin{document}
\twocolumn[{%
\renewcommand\twocolumn[1][]{#1}%
\maketitle
\begin{center}
\includegraphics[width=\textwidth]{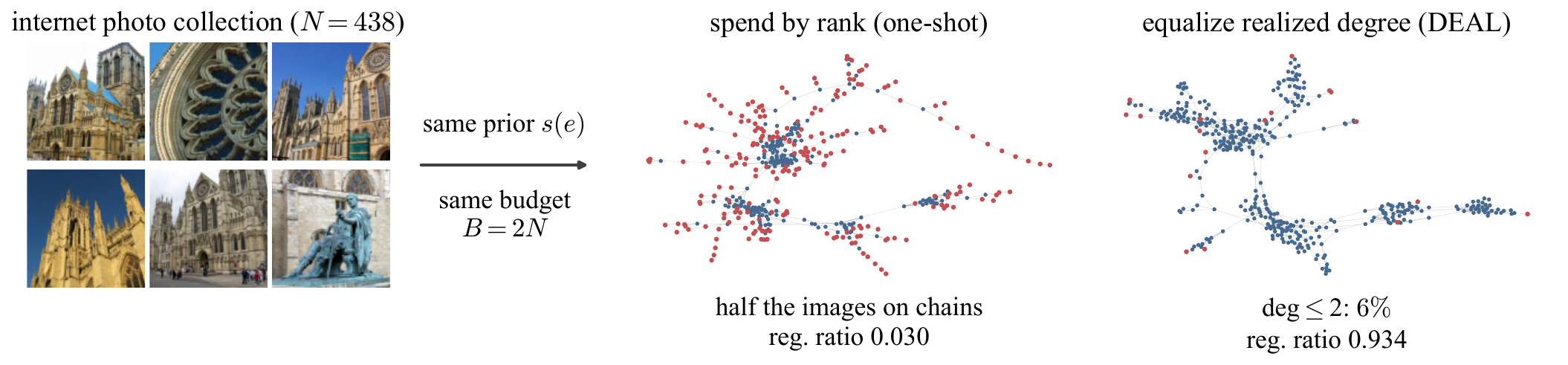}
\captionof{figure}{Yorkminster at $B{=}2N$ with the same prior. Top-$B$
allocation concentrates verified edges and registers $3\%$ of images. DEAL
spreads them across under-covered images and registers $93\%$. Red marks
degree at most two.}
\label{fig:teaser}
\end{center}
\vspace{2mm}
}]

\begingroup
\renewcommand{\endabstract}{\par}
\input{sec/00_abstract}
\endgroup
\newpage
\input{sec/01_intro}
\input{sec/02_related}
\input{sec/03_analysis}
\input{sec/04_method}
\input{sec/05_experiments}

\input{sec/06_discussion}
\input{sec/07_limitations}
\input{sec/07_conclusion}
\input{sec/08_acknowledgments}

\clearpage
{
    \small
    \bibliographystyle{ieeenat_fullname}
    \bibliography{refs}
}

\end{document}

%% file: authors.tex
\newcommand{\DEALAuthorBlock}{\begin{minipage}{\dimexpr\textwidth-2\tabcolsep\relax}
\centering
Minhyeok Im$^{1}$ \qquad Jongseong Brad Choi$^{2,3,4,*}$\par
\smallskip
{\small
$^{1}$Department of Computer Science, State University of New York, Stony Brook, NY, United States\par
$^{2}$Department of Mechanical Engineering, State University of New York, Korea, Incheon, South Korea\par
$^{3}$Department of Computer Science, State University of New York, Korea, Incheon, South Korea\par
$^{4}$Department of Mechanical Engineering, State University of New York, Stony Brook, NY, United States\par
\smallskip
$^{*}$Corresponding author\par}
\end{minipage}}
\author{\DEALAuthorBlock}

%% file: sec/00_abstract.tex
\begin{abstract}
Fully matching all image pairs in Structure-from-Motion is quadratic in the
number of images, so practical pipelines pass a budgeted subset to the final
matching and reconstruction backend. One-shot selectors score candidates
with image retrieval, cheap geometric probes, or a learned prior and finalize
$B$ pairs before full matching. Their priorities do not account for the degree
already allocated to each image, which can concentrate edges on a few images.
Across nine large 1DSfM internet photo collections at $B{=}2N$, the coverage
top-$B$ selector leaves $54\%$ of covered images with realized degree at most
two. Reconstruction can then fail even when the selected graph is nearly
connected and almost fully verified. Our training-free greedy rule,
Degree-Equalized Allocation (DEAL), exponentially downweights pairs whose
endpoints are already well covered. On these collections, DEAL improves the
one-shot selector built on the same prior ($0.776\!\to\!0.929$) and matches
the mean registered-image ratio of an SfM-supervised edge prior under the
shared backend ($0.929$ vs $0.930$), without training. On the largely
saturated 38-scene standard benchmark suite, DEAL gives the highest mean
among the evaluated training-free selectors ($0.875$ vs $0.858$ one-shot).
\end{abstract}

%% file: sec/01_intro.tex
\section{Introduction}
\label{sec:intro}

Incremental Structure-from-Motion (SfM)~\cite{schoenberger2016sfm,snavely2008photo}
reconstructs camera poses and sparse geometry from an image collection.
Fully matching all $\binom{N}{2}$ pairs is
quadratic in the number of images $N$, so practical pipelines pass only a
sparse subset to the final matching and reconstruction backend. Under a hard
budget of $B$ pairs, this selection bounds all downstream stages. An image left
disconnected by the graph cannot be registered. Setting $B{=}2N$ enforces a
strictly linear final-matching budget of two pair matches per input image, the
same $K{\cdot}N$ form with small $K$ that interleaved
systems have budgeted~\cite{lou2012matchminer}. The budget exceeds the
$N{-}1$ pairs of a spanning tree but leaves little slack for redundant
geometric constraints.

A selector chooses image pairs from a candidate pool using a scalar score,
which we call the prior. Full matching establishes correspondences for the
chosen pairs, geometric verification filters them, and the mapper estimates
camera poses and 3D points. Selection thus determines which image pairs can
provide geometric constraints to the reconstruction.

One-shot selectors differ in their scoring and repair rules but finalize the
budget before final matching.
Retrieval selectors rank pairs by appearance similarity, from vocabulary
trees~\cite{nister2006vocab} to NetVLAD~\cite{arandjelovic2016netvlad} and
DINOv2 neighbors~\cite{oquab2023dinov2}. Probe-scored selectors run a cheap
geometric check per candidate~\cite{wu2013linear,schoenberger2015paige}. 
Learned edge priors score pairs with a graph network trained on SfM outcomes~\cite{wei2026edgeprior}.
Although their scoring signals differ, these selectors compute pair priorities
before final matching and do not update them according to the degree already
allocated to each image. Consequently, the allocation can concentrate edges on
images favored by the prior. Across the nine 1DSfM collections evaluated
below at $B{=}2N$, the coverage top-$B$ selector leaves $54\%$ of covered images with
realized degree at most two, and incremental mappers can fail on the resulting
hub-and-chain graphs (\cref{fig:teaser}). Such failures need not appear as disconnected
graphs or as failed matches, since a selector can produce a nearly connected graph of
individually verified pairs whose reconstruction still fails (\cref{sec:analysis}).

We allocate the budget to reduce degree imbalance, favoring pairs that reach
under-covered images even when their prior score is lower.
Our greedy rule, Degree-Equalized Allocation (DEAL, \cref{sec:method}),
conditions each pick on evolving selected degree, with a bonus for
component-merging edges and an exponential decay for edges whose endpoints are
already well covered. Its observed form uses thresholded full-match counts
between rounds. Ablations attribute most of the gain to degree-aware allocation
rather than this feedback. We consider a fixed tight budget of $B{=}2N$ and evaluate downstream
reconstruction quality (\cref{sec:experiments:setup}).

The four standard benchmarks are largely saturated at $B{=}2N$. The one-shot baseline already registers
every image on 17 of 38 scenes. We therefore add nine larger internet photo collections from 1DSfM~\cite{wilson2014onedsfm},
where budgeted pair selection is more challenging. On these scenes, DEAL improves the coverage
top-$B$ selector from $0.776$ to $0.929$ in mean registered-image ratio and reaches the same
end-to-end ratio as a learned edge-prior system under the shared matching and mapping backend
($0.929$ vs. $0.930$). DEAL also reaches higher completeness-weighted rotation
mAA ($0.117$ vs. $0.098$), which combines registration and pose error.
Across the full 38-scene suite, DEAL reaches $0.875$, while the learned prior
remains slightly better overall at $0.919$.

We make two contributions.
\begin{enumerate}[label=(\arabic*), leftmargin=2.2em, itemsep=2pt, topsep=3pt, parsep=0pt]
\item \textbf{DEAL, a training-free degree-equalizing selection rule} for
budgeted view graphs, which contains one-shot top-$B$ selection as the
special case with its graph terms disabled. Ablations attribute most of the
gain to degree control rather than to the specific decay form
(\cref{sec:method}, \cref{sec:experiments}).
\item \textbf{A diagnosis of one-shot selection at tight budgets}, showing
that reconstruction can still fail with a nearly connected and almost fully
verified graph under severe degree concentration, together with graph-structure
and budget evaluations on nine large 1DSfM collections
(\cref{sec:analysis}, \cref{sec:experiments}).
\end{enumerate}

%% file: sec/02_related.tex
\section{Related Work}
\label{sec:related}

\noindent\textbf{View graphs in Structure-from-Motion.}
Incremental SfM builds a model from a verified view
graph~\cite{snavely2006phototourism,snavely2008photo,schoenberger2016sfm},
and global pipelines take the same graph as
input~\cite{sweeney2015theia,pan2024glomap}. DEAL operates at the
pre-matching stage that produces it and is agnostic to the downstream
solver.

\noindent\textbf{One-shot pair selection.}
The standard cheap selector ranks pairs by appearance. Examples are
vocabulary trees~\cite{nister2006vocab}, NetVLAD~\cite{arandjelovic2016netvlad},
GeM~\cite{radenovic2019gem}, DINOv2 neighbors~\cite{oquab2023dinov2}, and
retrieval inside hierarchical localization~\cite{sarlin2019coarse}.
MASt3R-SfM builds its graph from foundation-model features in one
shot~\cite{duisterhof2025mast3rsfm}. PAIGE learns a static pair-overlap
classifier~\cite{schoenberger2015paige}, and a recent learned edge prior
scores pairs with a graph network trained on SfM-derived supervision,
modulating scores by predicted connectivity and allocating over predicted
spanning structures~\cite{wei2026edgeprior}. All
of these finalize all $B$ pairs before full matching. The learned
prior's connectivity modulation adapts to a predicted topology, while DEAL
recomputes priorities from the evolving selected-degree state. Its observed
form can use full-match outcomes rather than assume every spent edge succeeds.

\noindent\textbf{Interleaved selection and verification.} Building Rome
in a Day alternates rounds of query expansion with geometric
verification~\cite{agarwal2009rome,chum2010largescale}. Image Webs prioritizes
component-merging pairs and re-ranks candidates by the Fiedler vector of the
revealed graph under an explicit verification budget~\cite{heath2010imagewebs}.
MatchMiner runs rounds under a $K{\cdot}N$ pair budget, feeds successes and
failures back into its shortlists, and weighs pairs by expected
component-entropy reduction~\cite{lou2012matchminer}. Kim et al.\ alternate
link prediction and verification with an explicit failure
matrix~\cite{kim2012matchgraph}, and GraphMatch samples and propagates from
verified edges until per-node saturation~\cite{cui2017graphmatch}. These
systems target graph completeness or component discovery. DEAL targets
reconstruction quality under a fixed near-$2N$ budget, and
\cref{sec:experiments:ablation} evaluates their objectives within the same
selection loop.

\noindent\textbf{View-graph sparsification and repair.}
A complementary line operates on a graph that already exists.
Skeletal graphs select a quality-preserving subset of an existing
match graph~\cite{snavely2008skeletal}. Viewing-graph optimization refines
fundamental matrices through loop consistency~\cite{sweeney2015viewing}.
Other methods filter inconsistent epipolar geometries and disambiguate
look-alike or duplicate structures~\cite{wilson2014onedsfm,zach2010loop,shen2016consistent,xiao2021progressive,heinly2014duplicate,cai2023doppelgangers,zhou2026pgvs,shah2017unified,cui2021viewgraph}.
SLAM selects pose-graph constraints under budgets by connectivity
criteria~\cite{tian2018nearoptimal,khosoussi2019reliable,denniston2022loopclosure},
including direct algebraic-connectivity
maximization~\cite{doherty2023mac}, but over realized or odometry-backed
edges. These methods consume matches as input, while budgeted selection
decides which matches will exist at all.

\noindent\textbf{Coverage as a conditioning signal.}
The spatial spread of correspondences is a long-standing conditioning
signal~\cite{hartley2003multiview,fraser1984network}, used by COLMAP for
next-image ranking~\cite{schoenberger2016sfm}. Our coverage prior applies
the same principle and scores probe inliers on a coverage grid. Learned
features and
matchers~\cite{lowe2004sift,detone2018superpoint,zhao2023aliked,sarlin2020superglue,lindenberger2023lightglue}
change which pairs survive, so every selector in our comparison runs through
one fixed backend (\cref{sec:experiments:setup}).

%% file: sec/03_analysis.tex
\section{Motivating Analysis}
\label{sec:analysis}
\label{sec:analysis:decomposition}

\begin{table}[H]
\caption{Selection quality at $B{=}2N$, as mean registered-image ratio over 38 scenes. Rows below the midrule alter multiple factors.}
\label{tab:decomp}
\vskip 0.1in
\begin{center}
\begin{footnotesize}\setlength{\tabcolsep}{3pt}
\begin{tabular}{@{}lcc@{}}
\toprule
selector & adds & mean ratio \\
\midrule
\dinotopb\ top-$B$        & appearance ranking      & $0.525$ \\
$+$ max-similarity MST    & connectedness           & $0.710$ \\
$+$ mid-rank band fill    & rank statistics         & $0.727$ \\
\midrule
\vocabtree\ (SIFT-verified) & geometric verification & $0.819$ \\
coverage top-$B$ & geometric probe scoring & $0.858$ \\
\wei\ (learned prior)     & learned global context  & $0.919$ \\
\bottomrule
\end{tabular}\end{footnotesize}
\end{center}
\vskip -0.1in
\end{table}

We first compare selection factors on 38 scenes from DTU, ETH3D,
Tanks\&Temples and IMC2023. Each selector spends $B{=}2N$ full pair matches
with the same ALIKED features, LightGlue matcher and COLMAP incremental
mapper. Quality is the fraction of input images registered in one model,
with zero for a failed reconstruction. \Cref{tab:decomp} compares selectors
under this protocol. Its first three rows form a controlled progression over
the same candidate pool. The last three change several factors together and
do not isolate individual contributions.

Within the controlled progression, graph connectedness gives the largest
observed increase (${+}0.185$) in registered-image ratio. Forcing a spanning
backbone over the same appearance ranking recovers $56\%$ of the
gap between retrieval methods and probe methods without additional matching.
Filling the remaining budget from moderate similarity ranks adds only
${+}0.017$, so rank position explains little beyond the backbone. A vocabulary
tree with SIFT-verified candidates reaches $0.819$, and the coverage top-$B$
selector reaches $0.858$ after probe-scoring every candidate with RANSAC. The
learned prior reaches $0.919$, though that comparison changes more than one
factor. These comparisons associate connectedness and verified geometry
with better registration, but a nearly connected graph of verified pairs can
still fail to reconstruct.

Yorkminster, one of the nine larger 1DSfM collections evaluated in
\cref{sec:experiments:unordered}, illustrates this remaining failure.
Coverage top-$B$ already includes connectivity repair and produces $2.6$
components on average across these collections. On Yorkminster, $866$ of its
$876$ selected pairs reach at least $15$ inliers, yet the mapper registers
only $13$ of $438$ images. The verified edges are also distributed unevenly across
images. Roughly half of the images hold at most two
of them while $15\%$ hold eight or more, so the graph couples a well-covered
core to long weakly redundant chains (\cref{fig:teaser}).

The result does not vary monotonically with the budget. The same
selector registers $0.904$ of Yorkminster at $B{=}1.5N$ and $0.920$ at $4N$
but only $0.030$ at $2N$ (\cref{sec:experiments:budget}). This non-monotonic
case is consistent with a mapper-sensitive regime in which additional
high-ranked edges alter the incremental trajectory without resolving weakly
supported chains. Selection at these budgets should account for how pairs are
allocated across images as well as how they are ranked. \Cref{sec:method}
introduces a rule for this allocation.

%% file: sec/04_method.tex
\section{Method}
\label{sec:method}

To reduce the degree concentration observed in \cref{sec:analysis}, DEAL
combines prior scores with the graph produced by earlier picks to favor
under-covered images. Its graph tracks image pairs, degrees and components.
The backend verifies full-match correspondences and passes them to the mapper,
while the allocation rule uses graph state rather than relative-pose
constraints.
\input{fig_overview}

\subsection{Problem Statement}
\label{sec:method:problem}

Given $N$ images, a candidate pool $\mathcal{C}$, a positive scalar prior
$s(e)$ per pair and budget $B$, the reconstruction objective is
\begin{equation}
\mathcal{E}^\star = \arg\max_{\mathcal{E}\subseteq\mathcal{C},\,
|\mathcal{E}|=B} R(\mathcal{E}),
\label{eq:objective}
\end{equation}
where $R$ is the fraction of input images registered in one reconstructed
model. Evaluating $R$ requires matching and mapping, so it is unavailable
during pair allocation. Selectors instead use prior scores as a surrogate.
These scores rank pairs rather than estimate calibrated success probabilities.
Pure top-$B$ selection treats their utility as additive and maximizes
\begin{equation}
\mathcal{E}_{\text{top}} = \arg\max_{|\mathcal{E}|=B}
\sum_{e\in\mathcal{E}} s(e).
\label{eq:oneshot}
\end{equation}
Sorting solves this additive objective without considering what earlier
pairs already cover. DEAL instead carries a graph $G_t$ with a degree
$d_t(u)$ for each image and updates pair priorities after every pick.
$G_t$ contains selected pairs under assumed success, or pairs passing the
observation threshold under observed updates.

\subsection{Degree-Equalized Allocation}
\label{sec:method:deal}

\noindent\textbf{Graph redundancy.}
For a connected graph of $N$ images and a subset $S$ of $k\le N/2$ images,
the variational form of algebraic connectivity gives
\begin{equation}
\lambda_2 \le \frac{N\,|\partial S|}{k\,(N-k)},
\label{eq:cheeger}
\end{equation}
where $\partial S$ contains the edges leaving $S$~\cite{cheeger1970,mohar1989isoperimetric}.
A chain attached through two edges has an upper bound of
\mbox{$2N/[k(N-k)]$}, so a large weakly attached subset can limit redundancy even
when the graph is connected. Pair ranking alone does not account for the
support already assigned to each endpoint. DEAL uses endpoint degree to
redirect spending toward under-covered images through the score below.

\noindent\textbf{Selection rule.}
In each round DEAL takes the unspent candidate with the highest gain
(\cref{fig:overview})
\begin{equation}
g(e) =
\begin{cases}
\beta \, s(e) & u \nsim_t v\\[2pt]
s(e)\, e^{-\gamma\, d^{\min}_t(e)} & \text{otherwise,}
\end{cases}
\label{eq:gain}
\end{equation}
where $u\sim_t v$ denotes membership in the same component and
$d^{\min}_t(e)=\min(d_t(u),d_t(v))$. Component-merging pairs receive a
multiplier $\beta$ without degree decay. Other pairs are downweighted when
both endpoints already have support. The minimum preserves the prior score
when either endpoint is uncovered. Unlike a hard threshold, the decay still
allows a high-scoring pair between covered images to outrank a low-scoring
one. Setting $\beta{=}1$ and $\gamma{=}0$ removes both graph terms and
recovers pure top-$B$ selection in \cref{eq:oneshot}.

\input{tab_main}
\noindent\textbf{Graph updates.}
The assumed-success form adds every spent pair to $G_t$. The observed form
adds a pair only when full LightGlue matching returns at least $\tau$
correspondences. This pre-MAGSAC count supplies feedback for the next round,
while the backend separately verifies pair geometry with a fundamental-matrix
fit. Observation adds no full matching because the spent pairs are the final
matches. The two update forms nearly coincide when selected pairs match at
high rates, as on 1DSfM (\cref{sec:experiments:ablation}).

\noindent\textbf{Priors.}
Both instantiations use a DINOv2 candidate pool. The DINOv2 prior is cosine
similarity. The coverage prior probes candidates with mutual nearest-neighbor
matching and RANSAC over the backend's ALIKED descriptors, then scores the
fraction of $4\times4$ grid cells containing an inlier. The corresponding
one-shot baselines use the same priors, isolating allocation from scoring.

\noindent\textbf{Cost.}
\label{sec:experiments:cost}
The loop rescans candidates after each pick and adds a dependency between
rounds, but no extra full matching. A batch is planned assuming its picks
succeed, then outcomes update $G$. With a resident LightGlue matcher,
Temple's sequential $604$-pair loop takes $12.7$\,s, including $10.1$\,s of
matching. Coverage probing over the $8.3$k-pair pool adds about $8$\,s outside
$B$. Batch sizes $8$ and $32$ give Jaccard $\ge0.997$ and identical
reconstructions. The sequential run reproduces the precomputed-outcome
simulation's selections exactly.

\begin{algorithm}[H]
\caption{DEAL: degree-equalized view-graph selection}
\label{alg:rule}
\begin{algorithmic}[1]
\REQUIRE candidates $\mathcal{C}$ with prior $s(e)$, budget $B$, batch $k$,
$\beta$, $\gamma$, threshold $\tau$
\STATE $G \leftarrow$ empty graph over the $N$ images
\WHILE{fewer than $B$ pairs spent}
  \STATE $P \leftarrow$ working copy of $G$
  \FOR{$i = 1, \dots, \min(k, \text{remaining budget})$}
    \STATE pick unspent $e^\ast = \arg\max_{e} g(e)$ by
    \cref{eq:gain} evaluated on $P$
    \STATE add $e^\ast$ to $P$ \COMMENT{in-round picks assume success}
  \ENDFOR
  \STATE match the picked batch \COMMENT{these matches are final}
  \STATE \textbf{observed form} add to $G$ pairs with at least $\tau$ full
  LightGlue matches, \textbf{assumed form} add the whole batch
\ENDWHILE
\STATE \textbf{return} the $B$ spent pairs and their matches
\end{algorithmic}
\end{algorithm}

%% file: fig_overview.tex
\begin{figure*}[t]
\centering
\includegraphics[width=0.9\textwidth]{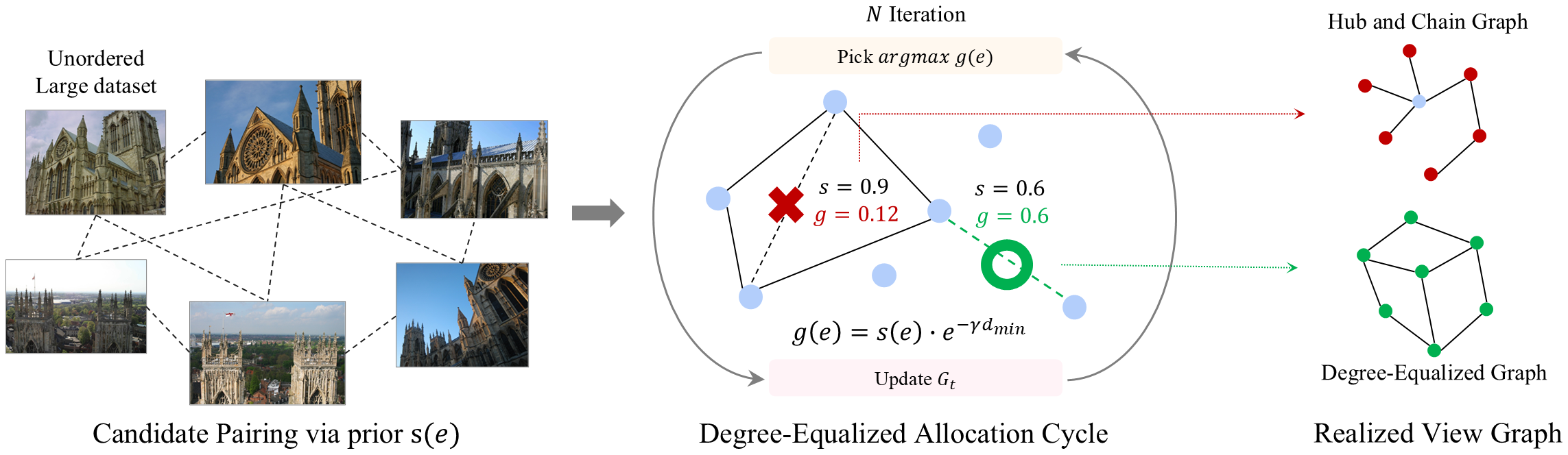}
\caption{DEAL overview. The prior scores each candidate pair once. DEAL
selects by \cref{eq:gain}, matches the pick, and updates the graph. Degree
decay lets an under-covered pair ($s{=}0.6$) outrank a covered pair
($s{=}0.9$). Red marks degree at most two.}
\label{fig:overview}
\end{figure*}

%% file: tab_main.tex
\begin{table*}[t]
\caption{Mean registered-image ratio on 38 scenes at $B{=}2N$. Rows above the divider allocate by rank. `fail' counts scenes below $0.1$, and bold marks the best training-free selector.}
\label{tab:main}
\vskip 0.1in
\begin{center}
\begin{small}\setlength{\tabcolsep}{4pt}\begin{tabular}{lcccccccc}
\toprule
Method & DTU & ETH3D & T\&T & IMC & ALL & ALL$\setminus$IMC & fail \\
\midrule
\dinotopb\ top-$B$ & 0.738 & 0.426 & 0.285 & 0.486 & 0.525 & 0.534 & 2/38 \\
\netvlad\ top-$B$ & 0.818 & 0.585 & 0.292 & 0.457 & 0.603 & 0.635 & 1/38 \\
\vocabtree & 0.986 & 0.730 & 0.966 & 0.573 & 0.819 & 0.875 & 4/38 \\
coverage top-$B$ & 0.995 & 0.839 & 0.891 & 0.612 & 0.858 & 0.913 & 1/38 \\
\wei~\cite{wei2026edgeprior} & 0.997 & 0.863 & 0.997 & 0.820 & 0.919 & 0.941 & 0/38 \\
\midrule
DEAL (DINOv2 prior) & 1.000 & 0.757 & 0.979 & 0.563 & 0.834 & 0.895 & 2/38 \\
DEAL (coverage prior) & 0.995 & 0.865 & 0.990 & 0.589 & \textbf{0.875} & 0.940 & 2/38 \\
\bottomrule
\end{tabular}\end{small}
\end{center}
\vskip -0.1in
\end{table*}

%% file: sec/05_experiments.tex
\section{Experiments}
\label{sec:experiments}

\subsection{Setup}
\label{sec:experiments:setup}

\noindent\textbf{Benchmarks.}
The benchmark suite covers 38 scenes drawn from DTU~\cite{jensen2014dtu} (13),
ETH3D~\cite{schoeps2017eth3d} (13), Tanks~\&~Temples~\cite{knapitsch2017tnt} (5),
and IMC2023~\cite{jin2021imc} (7). This suite is largely saturated at $B{=}2N$,
as the coverage top-$B$ selector already registers every image on 17 of its
38 scenes, so we add nine larger internet photo collections from
1DSfM~\cite{wilson2014onedsfm} under the same protocol.

\noindent\textbf{Protocol and metrics.}
Each selector receives a hard budget of $B{=}2N$ pairs, and its pairs are
matched and reconstructed by the same ALIKED+LightGlue+COLMAP incremental
backend. The backend initializes and refines a shared SIMPLE\_PINHOLE camera
without ground-truth intrinsics. The supplementary specifies this calibration
setup and also evaluates the same databases with the global GLOMAP solver.
The budget counts pairs passed to full LightGlue matching and reconstruction.
The coverage prior additionally performs a cheaper mutual-nearest-neighbor and
RANSAC probe over the candidate pool. This scoring cost lies outside $B$ and is
reported separately in \cref{sec:experiments:cost}. The
DINOv2 candidate pool is shared by every pool-based selector, while
\vocabtree\ and the learned prior generate their own. We report the
registered-image ratio, the fraction of images COLMAP registers into one
model, where $0.000$ means no model. COLMAP's mapper is nondeterministic, so
the DEAL configurations, the coverage top-$B$ selector and all 1DSfM numbers
are medians over three mapper repeats, and a difference of at most $0.01$ on
unrounded values counts as a tie.

\noindent\textbf{Baselines.}
Static baselines are \netvlad\ top-$B$, \dinotopb\ top-$B$, \vocabtree\ and
the learned edge prior of \wei~\cite{wei2026edgeprior}, run zero-shot from
its released IMC2023-trained checkpoint with its pairs fed through the
shared backend. Coverage top-$B$ starts from the $B$ highest coverage-ranked
pairs and swaps low-score cycle edges for component-merging candidates. It is
top-$B$ with connectivity repair, not the pure solution of \cref{eq:oneshot},
and is the direct one-shot reference because DEAL uses the same pool and prior. The two DEAL configurations
use the coverage prior and the DINOv2 prior with $\beta{=}5$, $\gamma{=}1$,
observation on, batch size $k{=}1$ and threshold $\tau{=}15$ pre-MAGSAC
LightGlue matches.

\subsection{Benchmark Suite at \texorpdfstring{$B{=}2N$}{B=2N}}
\label{sec:experiments:main}
\input{fig_graphs}

DEAL improves registration under the tight $B{=}2N$ budget (\cref{tab:main}). With the coverage prior,
it reaches $0.875$ on the 38-scene suite versus $0.858$ for coverage top-$B$ ($10/24/4$ W/T/L),
with the largest gains on scenes where the baseline remains below saturation
(wall $0.558{\to}1.000$, terrains $0.738{\to}1.000$, Temple $0.831{\to}0.997$).
Excluding IMC2023, where the benchmark is largely saturated and the coverage prior
is unreliable on several scenes, DEAL reduces the remaining registration error by
$31\%$ ($0.913{\to}0.940$). With the weaker DINOv2 prior, DEAL improves \dinotopb
from $0.525$ to $0.834$ without geometric probing, a $+0.309$ gain.

The IMC2023 results differ between the two priors. With the coverage prior,
DEAL matches or improves every family except IMC2023, where it falls to
$0.589$ versus $0.612$ for coverage top-$B$. The largest regressions occur on
the randomized-viewpoint scenes cyprus ($0.333{\to}0.000$), where DEAL produces
no model and raises the `fail' count from one to two, and dioscuri
($0.397{\to}0.328$). With the DINOv2 prior, equalization improves the IMC mean
($0.486{\to}0.563$). These opposite family-level trends indicate sensitivity
to the prior, although they do not identify the underlying cause of the
failures.

\wei\ has the highest suite mean at $0.919$, with most of its lead over DEAL
concentrated in IMC2023, the checkpoint's training family. Outside IMC, their
means are nearly identical ($0.940$ vs $0.941$). On chairs, the coverage
prior produces no model at $B{=}32$, while the DINOv2 prior reaches $0.750$,
further indicating sensitivity to the prior.

\subsection{Large Internet Collections}
\label{sec:experiments:unordered}
\input{tab_unordered}

On the nine 1DSfM scenes, DEAL closes the gap to the learned prior (\cref{tab:unordered}). 
It reaches $0.929$ versus $0.930$ for \wei. Under the unrounded $0.01$ tie
rule, it wins five scenes, ties three and loses Montreal. Against coverage top-$B$ it wins all nine
($0.776{\to}0.929$), and on Yorkminster it raises registration from $0.030$
to $0.934$ with the same pool, prior, and
budget, replacing the chained graph with an equalized one (\cref{fig:graphs}).
Excluding Yorkminster, DEAL still improves over coverage top-$B$ by $+0.059$ with eight wins
in eight, while remaining near parity with the learned prior ($0.928$ vs $0.931$).

The margin over coverage top-$B$ differs across the evaluated groups. It is
${+}0.018$ on the full suite (one-sided sign test $p{=}0.09$),
${+}0.076$ on the sixteen IMC+1DSfM scenes ($10/3/3$,
$p{=}0.046$) and ${+}0.153$ on the nine 1DSfM scenes (nine of nine,
$p{=}0.002$). On the sixteen IMC+1DSfM scenes, the learned prior retains the
highest mean ($0.882$, scene-wise $5/6/5$). Its advantage is concentrated in
IMC, its training family, and two failures of the coverage prior
(supplementary).

\subsection{Graph Structure}
\label{sec:experiments:mechanism}

The selected-graph statistics show the allocation difference
(\cref{tab:shape}). One-shot selectors concentrate edges on a small subset of images. Across
these selectors $34\%$ to $67\%$ of covered images have selected degree at most two,
compared with $6\%$ and $3\%$ for DEAL.

For all selectors except DINOv2 top-$B$, these fractions are nearly unchanged after matching
(within $0.03$), showing that the concentration is introduced during selection rather than 
by geometric verification. DINOv2 top-$B$ instead fails primarily through isolation,
with roughly half of the images receiving no verified edge. Its largest component holds
$26\%$ of the images, so the $\lambda_2$ of the full graph is zero and the value in
\cref{tab:shape} is not comparable to the other rows.
\input{tab_shape}

For coverage top-$B$ the realized graphs show the degree concentration
diagnosed in \cref{sec:analysis}.
Relative to coverage top-$B$, DEAL reduces the degree-$\le2$ fraction from
$54\%$ to $4\%$ and increases the algebraic connectivity of the largest
component by $2.5\times$ to $33\times$ across the nine 1DSfM scenes. The
component count does not improve ($3.1$ versus $2.6$), so DEAL does not
guarantee connectedness and $\lambda_2$ describes only the largest component.
These statistics diagnose fragile graphs rather than predict registration.
Degree imbalance falls broadly, while gains concentrate where the baseline
structure is particularly harmful.

Despite near-equal registration on 1DSfM, DEAL and the learned prior select
substantially different pair sets, with a mean IoU of $0.108$ (range
$0.077$--$0.154$), identical across repeats. Similar registration therefore
does not require close agreement between the selected pair sets.

\input{tab_ablation}

\subsection{Ablations}
\label{sec:experiments:ablation}

\noindent\textbf{Decay and hard-threshold ablation.}
Equalization, rather than the specific form of the penalty, accounts for most of the gain (\cref{tab:ablation}).
Removing the decay ($\gamma{=}0$) reduces the 1DSfM mean to $0.793$, Yorkminster to $0.027$
and the suite mean to $0.839$, where it wins no scene against the full configuration.
A one-batch hard-threshold variant uses $s(e)$ while
$d^{\min}{<}\kappa$ and $10^{-12}s(e)$ afterward for non-merging pairs. It
still rescans and fills $B$ under assumed success. At $\kappa{=}2$, the hard
threshold recovers most of the 1DSfM improvement ($0.925$, Yorkminster $0.927$)
but remains close to coverage top-$B$ on the full suite ($0.861$ vs
$0.858$). Increasing the threshold to $\kappa{=}4$ reduces the mean on both
sets ($0.897$ and $0.840$). Neither tested threshold matches the decay across
both evaluation sets, while a post-hoc $\gamma$ sweep stays within $0.012$ of
the default.

The suite mean is insensitive to the merging multiplier, but Montreal is not.
Setting $\beta{=}1$ leaves the suite mean at $0.875$ and raises the 1DSfM mean
from $0.929$ to $0.946$, almost entirely through Montreal
($0.736{\to}0.882$); $\beta{\in}\{2,10\}$ moves the mean by at most $0.001$.
We retain the fixed $\beta{=}5$ rather than tune it to this collection. The
multiplier is not the main source of the gain, but its scene-level effect is
not uniformly negligible.

Observing outcomes and assuming success produce identical selections on
eight of nine scenes and indistinguishable reconstructions ($0.929$ vs
$0.931$ on 1DSfM, $0.875$ vs $0.876$ on the suite). Only five selected pairs
miss the observation threshold, so full-match outcomes rarely change
subsequent scores in this setting.

Under assumed success, batch size does not affect the selection trajectory.
Setting $k{=}B$ gives a static planner that applies \cref{eq:gain} to selected
rather than realized degree. Plain top-$B$ is recovered only after also
setting $\beta{=}1$ and $\gamma{=}0$, which removes the graph-dependent gain.
Observation adds no matching cost and may matter more when selected pairs
fail more often.

Classical graph gains use the same loop. Entropy remains near coverage top-$B$
($0.786$). A Fiedler gain reaches $0.955$ on 1DSfM but requires an eigensolve
per pick and falls to $0.721$ on the sixteen IMC+1DSfM scenes and $0.821$ on
the suite, versus $0.780$ and $0.875$ for the decay (supplementary). Its higher
1DSfM mean therefore does not persist across the broader sets.

\subsection{Pose Accuracy}
\label{sec:experiments:pose}

Registration alone can hide misplaced cameras. On courtyard a contiguous
block of $15$ cameras is misplaced inside an accurate model while
every one of them counts as registered. \Cref{tab:pose} therefore reports
completeness-weighted mAA of rotation error over the thresholds
$\{1,2,3,5,10\}^\circ$ after RANSAC similarity alignment of the camera
centers, with the alignment rotation also applied to camera orientations.
Unregistered images fail every threshold, so mAA combines completeness and
pose accuracy. Median rotation error among registered cameras with reference
poses measures conditional accuracy, with camera support varying by selector.
The supplementary reports registration, mAA and this conditional error
separately for the pose-evaluation runs.

\input{tab_pose}

Pose follows registration on the suite, since completeness dominates the
metric and mAA equals the registration ratio on 11 of the 13 ETH3D scenes.
DEAL has the highest mAA on ETH3D ($0.823$ against $0.792$ for coverage
top-$B$ and $0.791$ for \wei), remains close to coverage top-$B$ overall
($0.770$ against $0.775$) and trails \wei\ ($0.807$). Its $0.321$ on IMC is
the lowest listed score (coverage top-$B$ reaches $0.414$), matching the
prior-sensitive regressions in \cref{sec:experiments:main}. On the 1DSfM collections the
reference is itself an SfM model and absolute values are low, so we compare
relative ordering. DEAL has the highest mAA ($0.117$ against $0.098$ for the
learned prior and $0.083$ for coverage top-$B$) and the lowest aggregate
median rotation error ($27^\circ$ against $30^\circ$ and $42^\circ$). Both
summaries favor DEAL on this collection set, while the per-scene ordering
varies (supplementary).

\newpage
\subsection{Budget Sweep}
\label{sec:experiments:budget}

The budget sweep bounds the regime in which degree equalization matters. On
the suite, the mean reverses at $4N$ ($0.898$ vs $0.932$, $4/30/4$ W/T/L) and
stays $0.010$ to $0.014$ behind coverage top-$B$ at $8N$ and $16N$ ($0.932$
vs $0.942$, $0.954$ vs $0.968$). The reversal does not appear on 1DSfM.
DEAL wins all nine scenes at $4N$ ($0.956$ vs $0.904$, $p{=}0.002$) and leads
at $1.5N$ ($0.923$ vs $0.875$, $7/2/0$, $p{=}0.008$; supplementary).
Coverage top-$B$ registers Yorkminster at both adjacent budgets ($0.904$ and
$0.920$), so its isolated $0.030$ at $2N$ is mapper-sensitive rather than a
general pair shortage. In \cref{fig:budget}, the learned prior converges with
the higher-performing selectors by $8N$, appearance retrieval reaches them by
$8N$ to $16N$, and the vocabulary tree stays near $0.8$.

\input{fig_budget}

%% file: fig_graphs.tex
\begin{figure*}[t]
\centering
\includegraphics[width=\textwidth]{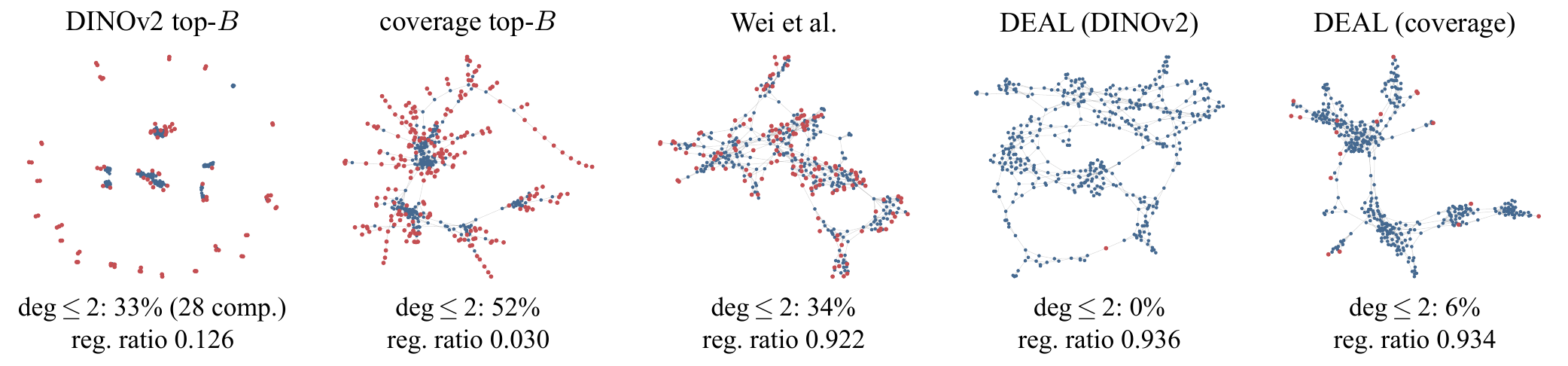}
\caption{Realized Yorkminster view graphs at $B{=}2N$. Red marks degree at
most two. DINOv2 top-$B$ shows the full graph, and the other panels show the
largest component. DEAL distributes verified edges more evenly than
rank-based selectors.}
\label{fig:graphs}
\end{figure*}

%% file: tab_unordered.tex
\begin{table}[b]
\caption{Registered-image ratio on nine 1DSfM collections at $B{=}2N$. Values are three-repeat medians except DINOv2 top-$B$, and W/T/L uses unrounded values with the $0.01$ tie rule. Bold marks the best per scene.}
\label{tab:unordered}
\vskip 0.1in
\begin{center}
\begin{footnotesize}\setlength{\tabcolsep}{2.0pt}\begin{tabular}{@{}lccccc@{}}
\toprule
 & \multicolumn{2}{c}{top-$B$} & learned & \multicolumn{2}{c}{DEAL} \\
\cmidrule(lr){2-3}\cmidrule(lr){4-4}\cmidrule(lr){5-6}
Scene & DINOv2 & coverage & \wei & DINOv2 & coverage \\
\midrule
Alamo & 0.323 & 0.938 & 0.947 & \textbf{0.950} & \textbf{0.950} \\
Ellis Island & 0.304 & 0.883 & 0.930 & 0.856 & \textbf{0.940} \\
Gendarmenmarkt & 0.195 & 0.919 & 0.947 & 0.956 & \textbf{0.971} \\
Madrid Metropolis & 0.302 & 0.940 & 0.947 & 0.933 & \textbf{0.956} \\
Montreal Notre Dame & 0.413 & 0.720 & \textbf{0.880} & 0.729 & 0.736 \\
NYC Library & 0.289 & 0.731 & 0.936 & 0.900 & \textbf{0.948} \\
Piazza del Popolo & 0.193 & 0.914 & 0.974 & 0.958 & \textbf{0.982} \\
Tower of London & 0.108 & 0.909 & 0.883 & 0.910 & \textbf{0.941} \\
Yorkminster & 0.126 & 0.030 & 0.922 & \textbf{0.936} & 0.934 \\
\midrule
mean & 0.250 & 0.776 & \textbf{0.930} & 0.903 & 0.929 \\
\bottomrule
\end{tabular}\end{footnotesize}
\end{center}
\vskip -0.1in
\end{table}

%% file: tab_shape.tex
\begin{table}[t]
\caption{Mean view-graph structure over nine 1DSfM scenes at $B{=}2N$. `selected' and `realized' report degree at most two before and after verification. Degree fractions use covered images, and $\lambda_2$ uses the largest component.}
\label{tab:shape}
\vskip 0.1in
\begin{center}
\begin{small}\setlength{\tabcolsep}{4pt}\begin{tabular}{lccccc}
\toprule
 & \multicolumn{2}{c}{deg$\le$2} & & & \\
\cmidrule(lr){2-3}
Selector & selected & realized & deg$\ge$8 & comp. & $\lambda_2$ \\
\midrule
\dinotopb\ top-$B$ & 0.67 & 0.35 & 0.39 & 26.8 & 0.100 \\
coverage top-$B$ & 0.54 & 0.54 & 0.15 & 2.6 & 0.007 \\
\wei & 0.34 & 0.35 & 0.08 & 3.2 & 0.027 \\
\midrule
DEAL (DINOv2) & 0.03 & 0.06 & 0.02 & 1.0 & 0.010 \\
DEAL (coverage) & 0.06 & 0.04 & 0.05 & 3.1 & 0.045 \\
\bottomrule
\end{tabular}\end{small}
\end{center}
\vskip -0.1in
\end{table}

%% file: tab_ablation.tex
\begin{table}[t]
\caption{Ablation at $B{=}2N$, as three-repeat median registered-image ratios. `Min' is the worst 1DSfM scene. Threshold controls use one assumed-success batch and still spend all $B$ pairs.}
\label{tab:ablation}
\vskip 0.1in
\begin{center}
\begin{small}\setlength{\tabcolsep}{3pt}\begin{tabular}{lccc}
\toprule
 & \multicolumn{2}{c}{1DSfM (9)} & suite (38) \\
\cmidrule(lr){2-3}\cmidrule(lr){4-4}
Variant & mean & min & mean \\
\midrule
coverage top-$B$ & 0.776 & 0.030 & 0.858 \\
\midrule
hard threshold $\kappa{=}4$ & 0.897 & 0.716 & 0.840 \\
hard threshold $\kappa{=}2$ & 0.925 & 0.731 & 0.861 \\
\midrule
DEAL, $\gamma{=}0$ (no decay) & 0.793 & 0.027 & 0.839 \\
DEAL, $\beta{=}1$ (no bridge multiplier) & 0.946 & 0.882 & 0.875 \\
DEAL, entropy gain (MatchMiner) & 0.786 & 0.027 & 0.828 \\
DEAL, Fiedler gain (Image Webs) & 0.955 & 0.934 & 0.821 \\
DEAL, assumed success & 0.931 & 0.738 & 0.876 \\
DEAL, observed (default) & 0.929 & 0.736 & 0.875 \\
\bottomrule
\end{tabular}\end{small}
\end{center}
\vskip -0.1in
\end{table}

%% file: tab_pose.tex
\begin{table}[t]
\caption{Completeness-weighted rotation mAA at $B{=}2N$ on 33 scenes with pose ground truth. Unregistered images count as failures. The five Tanks~\&~Temples Advanced scenes are excluded because their reference camera poses are not public. The IMC reference is a COLMAP model.}
\label{tab:pose}
\vskip 0.1in
\begin{center}
\begin{small}\setlength{\tabcolsep}{4pt}\begin{tabular}{lcccc}
\toprule
Selector & DTU & ETH3D & IMC & ALL \\
\midrule
\dinotopb & 0.718 & 0.420 & 0.379 & 0.529 \\
\netvlad & 0.785 & 0.584 & 0.324 & 0.608 \\
\vocabtree & 0.953 & 0.694 & 0.377 & 0.729 \\
coverage top-$B$ & 0.952 & 0.792 & 0.414 & 0.775 \\
\wei & 0.961 & 0.791 & 0.553 & 0.807 \\
\midrule
DEAL (DINOv2 prior) & 0.959 & 0.683 & 0.442 & 0.741 \\
DEAL (coverage prior) & 0.960 & 0.823 & 0.321 & 0.770 \\
\bottomrule
\end{tabular}\end{small}
\end{center}
\vskip -0.1in
\end{table}

%% file: fig_budget.tex
\begin{figure}[t]
\centering
\includegraphics[width=\columnwidth]{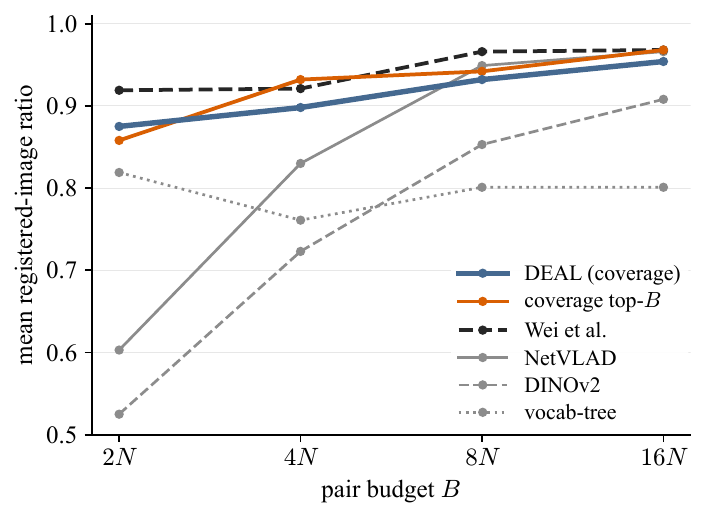}
\caption{Mean registered-image ratio over 38 scenes versus pair budget. DEAL
and the $2N$ coverage top-$B$ and \wei\ points are three-repeat medians. The
other points are single runs.}
\label{fig:budget}
\end{figure}

%% file: sec/06_discussion.tex
\section{Discussion}
\label{sec:discussion}

The ablations identify selected-degree regularization as the main source of
DEAL's gain. Degree fractions and $\lambda_2$ describe the structures it
changes, while the mapper trajectory determines whether these structures
cause failure. Observation remains available when selected pairs fail.

DEAL helps most when the prior proposes verifiable pairs but concentrates
the budget, as on 1DSfM. Its gains are less consistent on the IMC scenes,
where the observed gains differ between the two priors.

Coverage probing adds preprocessing but gives the higher DEAL mean. DINOv2
avoids that probe at lower suite accuracy ($0.834$). The learned prior retains
the highest overall mean, led by its IMC training family.

%% file: sec/07_limitations.tex
\section{Limitations}
\label{sec:limitations}

DEAL inherits omissions in its candidate pool and errors in its prior, and
does not detect an unreliable prior or switch allocations online. Evidence for its benefit is
concentrated near $B{\approx}2N$; on the 38-scene suite, its advantage over
coverage top-$B$ narrows or reverses as the budget grows.

The degree-based gain is an empirical surrogate for registration, with no
approximation guarantee for \cref{eq:objective}. Degree balance and
connectedness alone do not certify viewing-graph solvability, which also
depends on geometric constraints~\cite{trager2018solvability}. The backend's
pairwise verification filters correspondences but does not enforce multi-view
loop consistency. Such optimization is complementary to allocation.

The main evaluation uses one matcher and one incremental mapper. In the
supplementary GLOMAP evaluation over the same databases, DEAL retains the
highest mean ($0.946$, $8/1/0$ against the learned prior), while GLOMAP
partially recovers some weak coverage top-$B$ cases. This sensitivity
indicates that the magnitude of these failures depends partly on the mapper.
Registered-image ratio can
also overstate quality, as on courtyard. We therefore report pose metrics
where public reference poses are available and exclude the five
Tanks~\&~Temples Advanced scenes, whose reference camera poses are not public.

%% file: sec/07_conclusion.tex
\section{Conclusion}
\label{sec:conclusion}

Budgeted view-graph selection is commonly formulated as ranking, but
allocation across images also affects reconstruction under tight budgets.
On 1DSfM, DEAL raises the coverage top-$B$ mean from $0.776$ to $0.929$ and
matches the learned system under the shared backend. Relative to coverage
top-$B$, its realized graphs
reduce the mean degree-$\le2$ fraction from $54\%$ to $4\%$ and raise
algebraic connectivity by $2.5\times$ to $33\times$ on the same scenes. These
statistics diagnose weak chains rather than predict registration directly.

%% file: sec/08_acknowledgments.tex
\section*{Acknowledgments}
This work was supported by the Technology Innovation Program (Game-Changing Technology Project, No. RS-2026-25545225, Development of a Real-World Integrated Spatial Intelligence Platform Based on Multimodal 3D Space Generation and Hyper-Intelligent Operating Engines) funded by the Ministry of Trade, Industry and Energy (MOTIE), Korea.